\documentclass[runningheads]{llncs}
\pdfoutput=1
\usepackage{makeidx}
\usepackage{graphicx}
\usepackage{amsmath,amssymb} 
\usepackage{color}
\usepackage{multirow}
\usepackage{hyperref}
\usepackage{booktabs}

\begin{document}
	\pagestyle{headings}
	\mainmatter

	\title{Single Level Feature-to-Feature Forecasting
    \texorpdfstring{\\}{}
    with Deformable Convolutions}

	\titlerunning{}
	\authorrunning{}
\author{
Josip {\v{S}}ari{\'c}\inst{1} \and
Marin Or{\v{s}}i{\'c}\inst{1} \and
Ton{\'c}i Antunovi{\'c}\inst{2} \and
Sacha Vra{\v{z}}i{\'c}\inst{2} \and
Sini{\v{s}}a {\v{S}}egvi{\'c}\inst{1}
}

\institute{
$^1$ University of Zagreb, Faculty of Electrical Engineering and Computing, Croatia\\
$^2$ Rimac Automobili, Sveta Nedelja, Croatia\\
}

	\maketitle
\begin{abstract}
  Future anticipation is of vital importance
  in autonomous driving
  and other decision-making systems.
  We present a method to anticipate 
  semantic segmentation of future frames 
  in driving scenarios
  based on feature-to-feature forecasting.
  Our method is based on a 
  semantic segmentation model
  without lateral connections
  within the upsampling path.
  Such design ensures that
  the forecasting addresses 
  only the most abstract features
  on a very coarse resolution.
  We further propose to express
  feature-to-feature forecasting
  with deformable convolutions.
  This increases the modelling power
  due to being able to represent
  different motion patterns 
  within a single feature map. 
  Experiments show that our models
  with deformable convolutions 
  outperform their regular 
  and dilated counterparts
  while minimally increasing
  the number of parameters.
  Our method achieves 
  state of the art performance on 
  the Cityscapes validation set
  when forecasting
  nine timesteps into the future.
\end{abstract}

\section{Introduction}
Ability to anticipate the future is an 
important attribute of intelligent behavior,
especially in decision-making systems such as
robot navigation and autonomous driving.
It allows to plan actions 
not only by looking at the past,
but also by considering the future.
Accurate anticipation is critical
for reliable decision-making
of autonomous vehicles.
The farther the forecast,
the longer the time to avoid 
undesired outcomes of motion.
We believe that semantic forecasting  
will be one of critical concepts 
for avoiding accidents
in future autonomous driving systems.

There are three meaningful levels 
at which forecasting could be made: 
raw images, 
feature tensors, 
and semantic predictions. 
Forecasting raw images 
\cite{vukotic2017one,mathieu2015deep}
is known to be a hard problem.
Better results have been obtained
with direct forecasting 
of semantic segmentation predictions
\cite{luc2017predicting}. 
The third approach is to 
forecast feature tensors
instead of predictions 
\cite{vondrick2015anticipating}.
Recent work \cite{luc2018predicting}
proposes a bank of feature-to-feature (F2F) models
which target different resolutions
along the upsampling path of a 
feature pyramid network \cite{lin2017feature}.
Each F2F model receives corresponding features
from the four previous frames (t, t-3, t-6, t-9)
and forecasts the future features (t+3 or t+9).
The forecasted features are used 
to predict instance-level segmentations
\cite{he2017mask}
at the corresponding resolution 
level.

This paper addresses 
forecasting of future semantic segmentation maps
in road driving scenarios. 
We propose three improvements
with respect to the original
F2F approach \cite{luc2018predicting}.
Firstly, we base our work 
on a single-frame model 
without lateral connections. 
This requires only one F2F model 
which targets the final features 
of the convolutional backbone. 
These features are 
very well suited 
for the forecasting task 
due to high semantic content 
and coarse resolution.
Secondly, we express our F2F model 
with deformable convolutions
\cite{zhu2018deformable}.
This greatly increases the modelling power
due to capability to account
for different kinds of motion patterns 
within a single feature map. 
Thirdly, we provide an opportunity 
for the two independently trained submodels
(F2F, upsampling path)
to adapt to each other
by joint fine-tuning.
This would be very difficult to achieve
with multiple F2F models \cite{luc2018predicting}
since the required set of cached activations 
would not fit into GPU memory.
Thorough forecasting experiments 
on Cityscapes val \cite{cordts2016cityscapes} demonstrate 
state-of-the-art mid-term (t+9) performance
and runner-up short-term (t+3) performance
where we come second only to 
\cite{terwilliger2018recurrent}
who require a large computational effort 
to extract optical flow prior the forecast.
Two experiments on Cityscapes test 
suggest that our performance estimates 
on the validation subset
contain very little bias (if any).

\section{Related Work}
\label{sec:relatedwork}

\paragraph{Semantic segmentation.}
State of the art methods 
for semantic segmentation 
\cite{Zhao_2017_CVPR,chen2018deeplab,yang18cvpr,krevso2019efficient} 
have overcome the 80\% mIoU barrier
on Cityscapes test.
However, these methods are not well suited 
for F2F forecasting 
due to huge computational cost
and large GPU memory footprint. 
We therefore base our research
on a 
recent 
semantic segmentation model
\cite{orvsic2019defense}
which achieves a great ratio between 
accuracy (75.5 mIoU Cityscapes test) and
speed (39 Hz on GTX1080Ti with 2MP input).
This model is a great candidate for
F2F \cite{luc2018predicting} forecasting due to a backbone 
with low-dimensional features
(ResNet-18, 512D)
and a lean upsampling path 
similar to FPN \cite{lin2017feature}.
In particular, we rely on a 
slightly impaired version 
of that model (72.5 mIoU Cityscapes val) 
with no lateral connections in the upsampling path.

\paragraph{Raw image forecasting.}
Predicting future images is interesting  
because it opens opportunities 
for unsupervised representation learning 
on practically unlimited data. 
It has been studied in many directions: 
exploiting adversarial training \cite{mathieu2015deep} 
anticipating arbitrary future frames \cite{vukotic2017one}, 
or leveraging past forecasts 
to autoregressively anticipate 
further into the future
\cite{kalchbrenner2017video}.

\paragraph{Feature forecasting.}
Feature-level forecasting has been first used
to anticipate appearance and actions in video
\cite{vondrick2015anticipating}. 
The approach uses past features 
to forecast the last AlexNet layer
of a future frame.
Later work \cite{luc2018predicting}
forecasts convolutional features
and interprets them with the Mask-RCNN 
\cite{he2017mask} 
head of the single-frame model.
F2F approaches are applicable 
to dense prediction tasks 
such as
panoptic segmentation \cite{kirillov2018panoptic}, semantic segmentation \cite{Zhao_2017_CVPR}, optical flow \cite{sun2018pwc} etc. 

\paragraph{Semantic segmentation forecasting.}
Luc et al.\ \cite{luc2017predicting} 
set a baseline 
for direct semantic segmentation forecasting 
by processing softmax preactivations 
from past frames. 
Nabavi et al.\ \cite{rochan2018future} 
train an end-to-end model which
forecasts intermediate features by 
convolutional LSTM \cite{xingjian2015convolutional}. 
Bhattacharyya et al.\ \cite{bhattacharyya2018bayesian} 
use Bayesian learning to model 
the multi-modal nature of the future 
and directly predict 
future semantic segmentation 
of road driving scenes. 
None of the previously mentioned approaches 
utilize optical flow 
despite its usefulness for video recognition \cite{feichtenhofer16cvpr}. 
Jin et al.\ \cite{jin2017predicting} 
jointly forecast 
semantic segmentation predictions 
and optical flow. 
They use features 
from the optical flow subnet 
to provide better future semantic maps. 
Terwilliger et al.\ \cite{terwilliger2018recurrent}
predict future optical flow 
and obtain future prediction by 
warping the semantic segmentation map 
from the current frame. 

\paragraph{Convolutions with a wide field of view.}
Convolutional models \cite{lecun1995convolutional}
proved helpful in most visual recognition tasks.
However, stacking vanilla convolutional layers
often results in undersized receptive field.
Consequently, the receptive field has been enlarged
with dilated convolutions \cite{yu2015multi} 
and spatial pyramid pooling \cite{Zhao_2017_CVPR}.
However, these techniques are unable 
to efficiently model geometric warps 
required by F2F forecasting.
Early work on warping 
convolutional representations
involved a global affine transformation
at the tensor level \cite{jaderberg15nips}.
Deformable convolutions \cite{dai17iccv}
extend this idea by introducing 
per-activation convolutional warps
which makes them especially 
well-suited for F2F forecasting.

\section{Single-Level F2F model with Deformable Convolutions}

\begin{figure}[t]
    \centering
    \begin{tabular}{c@{\quad\qquad}c} 

    \includegraphics[height=5.3cm]{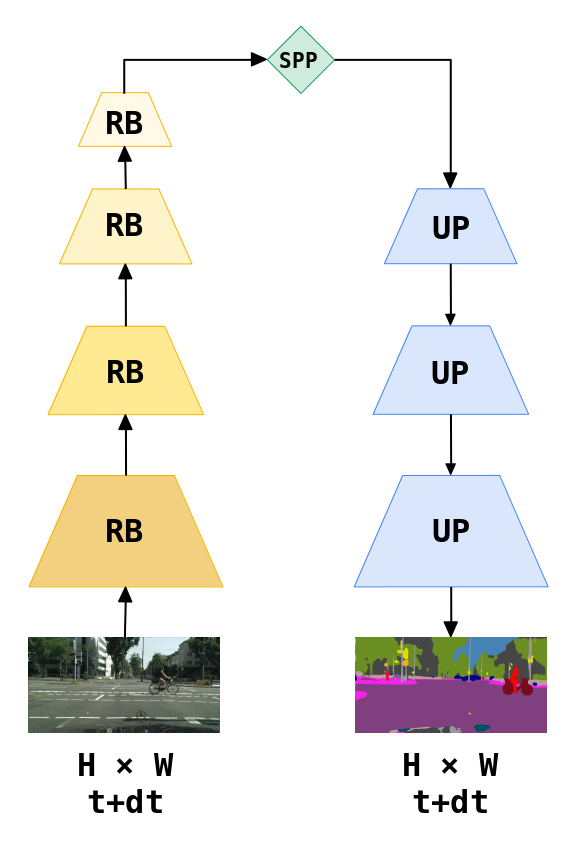}
    &
    \includegraphics[height=5.3cm]{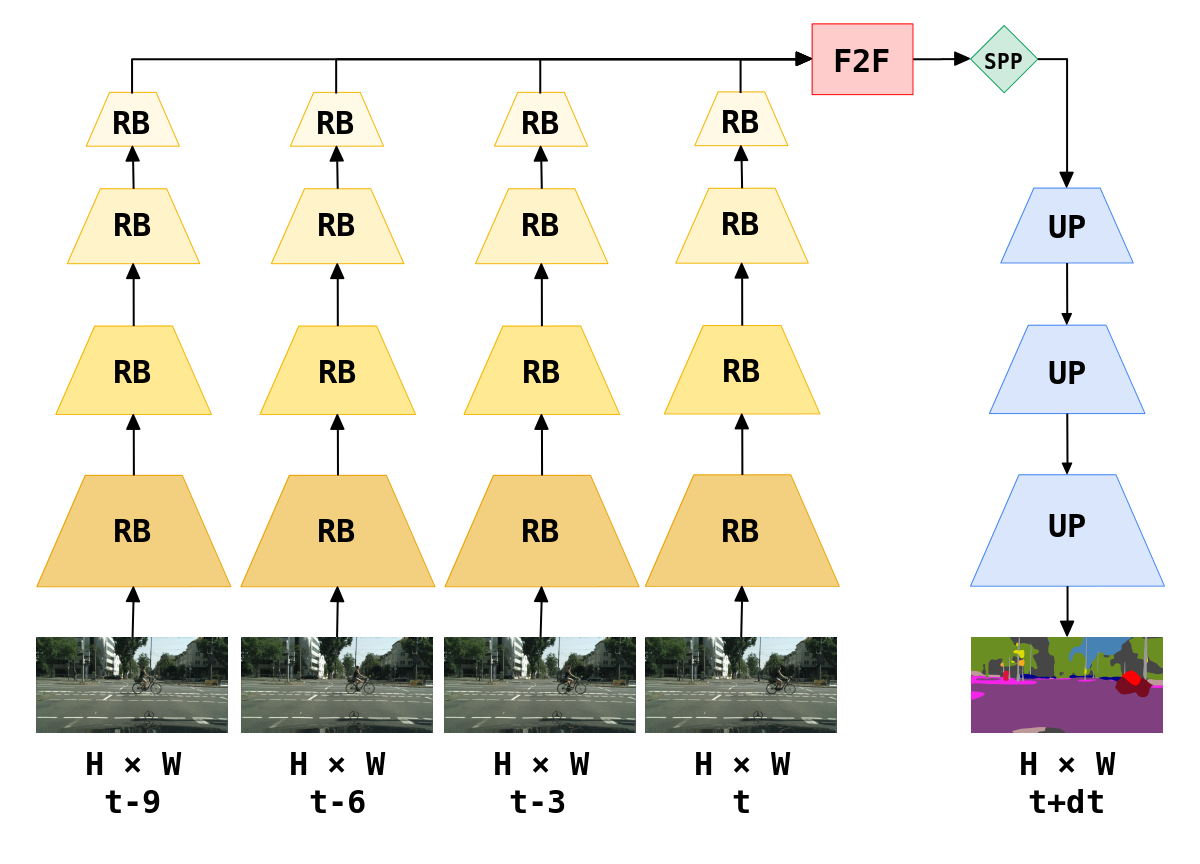}\\
    (a) & (b)
    \end{tabular}
    \caption{Structural diagram of 
      the employed single-frame model (a) 
      and the proposed compound model for 
      forecasting semantic segmentation (b).
      The two models share 
      the ResNet-18 feature extractor (yellow)
      and the upsampling path (green, blue).
    }
    \label{fig:model}
\end{figure}

We propose a method for semantic segmentation 
forecasting composed of
i) feature extractor (ResNet-18),
ii) F2F forecasting model,
and iii) upsampling path,
as illustrated in Fig.\,~\ref{fig:model} (b).
Yellow trapezoids represent
ResNet processing blocks RB1 - RB4
which form the feature extractor.
The red rectangle represents the F2F model.
The green rhombus designates 
spatial pyramid pooling (SPP) 
while the blue trapezoids 
designate modules which form 
the upsampling path.

Fig.\,\ref{fig:model}(a) shows 
the single-frame model 
which we use to train 
the feature extractor
and the upsampling path.
We also use this model as an oracle 
which predicts future segmentation
by observing a future frame.
Experiments with the oracle 
estimate upper performance bound
of semantic segmentation forecasting.

\subsection{Training Procedure}

The training starts from a public parameterization
of the feature extractor pre-trained on ImageNet \cite{deng2009imagenet}.
We jointly train the feature extractor
and the upsampling path for single-frame
semantic segmentation \cite{orvsic2019defense}.
We use that model to extract features 
at times t-9, t-6, t-3, and t (sources),
as well as at time t+dt (target).
We then train the F2F model 
with L2 loss in an unsupervised manner.
However, the forecasting 
induces a covariate shift
due to imperfect F2F prediction.
Therefore, we adapt the upsampling path
to noisy forecasted features
by fine-tuning the F2F model
and the upsampling path 
using cross-entropy loss
with respect to ground truth labels.
We update the F2F parameters
by averaging gradients 
from  F2F L2 loss
and the backpropagated cross-entropy loss.

\subsection{Proposed Feature-to-Feature Model}

We propose a single-level F2F model
operating on features from the last 
convolutional layer of ResNet-18.
We formulate our model as 
a sequence of N deformable convolutions
and denote it as DeformF2F-N. 
The first convolution of the sequence 
has the largest number of input feature maps 
since it blends features from all previous frames. 
Therefore we set its kernel size to $1\times1$. 
All other convolutions have $3\times3$ kernels
and 128 feature maps, except the last one
which recovers the number of feature maps 
to match the backbone output.

The proposed formulation differs
from the original F2F architecture
\cite{luc2018predicting} 
in three important details.
Firstly, we forecast backbone features
instead of features from the upsampling path.
Backbone features have a larger dimensionality,
and are closer to ImageNet pre-trained parameters
due to reduced learning rate during joint training.
Hence, these features are more distinctive
than features trained for recognition of only 19 classes.
Forecasting SPP features decreased the validation performance
for 1 percentage point (pp) mIoU in early experiments.

Secondly, we use a single-level F2F model
which performs the forecasting at 
a very coarse resolution
(1/32 of the original image).
This is beneficial since 
small feature displacements
simplify motion prediction
(as in optical flow).
Early multi-level forecasting experiments 
decreased performance for 2 pp mIoU.

Thirdly, we use thin deformable convolutions \cite{dai17iccv} 
instead of thick dilated ones. 
This decreases the number of parameters
and improves the performance 
as presented in ablation experiments. 
Feature-to-feature forecasting
is rather geometrically
than semantically heavy,
since the inputs and the outputs
are at the same semantic level.
Regular convolutions lack the potential 
to learn geometrical transformations
due to fixed grid sampling locations.
In deformable convolutions, 
the grid sampling locations 
are displaced with learned 
per-pixel offsets which are inferred 
from the preceding feature maps. 
We believe that learnable displacements 
are a good match 
for F2F transformation 
since they are able to model 
semantically aware per-object dynamics
across observed frames.

\subsection{Inference}

The proposed method requires
features from four past frames.
These features are concatenated 
and fed to the F2F module
which forecasts the future features.
The future features are fed 
to the upsampling path which predicts 
the future semantic segmentation. 
A perfect F2F forecast would 
attain performance of the single-frame model 
applied to the future frame,
which we refer to as oracle.

The proposed method is suitable
for real-time semantic forecast
since the feature extractor 
needs to be applied only once per frame.
Consider the computational complexity
of the single-frame model as baseline.
Then the only overhead for a single forecast 
corresponds to caching of four feature tensors
evaluating the F2F model.
If we require both the current prediction and a single forecast,
then the overhead would additionally include 
one evaluation of the upsampling path.
\section{Experiments}
We perform experiments 
on the Cityscapes dataset \cite{cordts2016cityscapes} 
which contains 2975 training,
500 validation and
1525 test images
with dense labels from 19 classes. 
The dataset includes 19 preceding 
and 10 succeeding unlabeled 
frames for each image. 
Each such mini-clip is 1.8 seconds long. 
Let $\mathbf{X}$ denote features from 
the last convolutional layer of ResNet-18.
The shape of these features is 
512$\times$H/32$\times$W/32, 
where 512 is the number of feature maps, 
while H and W are image dimensions. 
Then, the model input 
is a tuple of features 
$\left(
\mathbf{X}_{t - 9},
 \mathbf{X}_{t - 6},
 \mathbf{X}_{t - 3},
 \mathbf{X}_{t}
 \right)
 $.
The model output are future features 
$\mathbf{X}_{t+3}$ (short-term prediction, 0.18\,s) or 
$\mathbf{X}_{t+9}$ (mid-term prediction, 0.54\,s)
\cite{luc2018predicting}
which in most experiments correspond 
to the labeled frame in a mini-clip. 

\subsection{Implementation Details}

We use the deformable convolution implementation 
from \cite{mmdetection2018}. 
The features are pre-computed from full-size
Cityscapes images and stored on SSD drive. 
We optimize the L2 regression loss with Adam \cite{kingma2014adam}. 
We set the learning rate to 5e-4 
and train our F2F models for 160 epochs 
with batch size 12 in all experiments. 
We fine-tune our model with SGD 
with learning rate set to 1e-4
and batch size 8
for 5 epochs.
The training takes around 6 hours 
on a single GTX1080Ti. 

We measure semantic segmentation performance 
on the Cityscapes val dataset. 
We report the standard mean intersection over union metric over all 19 classes. 
We also measure mIoU for 8 classes representing \textbf{m}oving \textbf{o}bjects (person, rider, car, truck, bus, train, motorcycle, and bicycle).

\subsection{Comparison with the State of the Art on Cityscapes Val}
\label{ss:qr}

Table \ref{tab:mainres} evaluates several models 
for semantic segmentation forecasting. 
The first section shows 
the performance of the oracle, and 
the copy-last-segmentation baseline
which applies the single-frame model 
to the last observed frame. 
The second section shows 
results from the literature. 
The third section shows our results. 
The last section shows our result
when F2F model is trained 
on two feature tuples per mini-clip.
The row Luc F2F applies the model 
proposed in \cite{luc2018predicting} 
as a component of our method. 
The methods DeformF2F-5 and DeformF2F-8 
correspond to our models with 5 and 8 
deformable convolutions respectively. 
The suffix FT denotes that 
our F2F model is fine-tuned 
with cross entropy loss.

\begin{table*}[h]
\caption{Semantic forecasting on the Cityscapes validation set.}
\begin{center}
{\small
\setlength{\tabcolsep}{6pt}
\begin{tabular}{lcccc}
\toprule
& \multicolumn{2}{c}{Short-term} & \multicolumn{2}{c}{Mid-term} \\
& mIoU & mIoU-MO & mIoU & mIoU-MO \\
\midrule
Oracle & 72.5 & 71.5 & 72.5 & 71.5 \\
Copy last segmentation & 52.2 & 48.3 & 38.6 & 29.6 \\
\midrule
Luc Dil10-S2S \cite{luc2017predicting} & 59.4 & 55.3 & 47.8 & 40.8 \\
Luc Mask-S2S \cite{luc2018predicting} & / & 55.3 & / & 42.4 \\
Luc Mask-F2F \cite{luc2018predicting} & / & 61.2 & / & 41.2 \\
Nabavi \cite{rochan2018future} & 60.0 & / & / & / \\
Terwilliger \cite{terwilliger2018recurrent} & \textbf{67.1} & \textbf{65.1} & 51.5 & 46.3 \\ 
Bhattacharyya \cite{bhattacharyya2018bayesian} & 65.1 & / & 51.2 & / \\
\midrule
Luc F2F (our implementation) & 59.8 & 56.7 & 45.6 & 39.0 \\
DeformF2F-5 & 63.4 & 61.5 & 50.9 & 46.4 \\
DeformF2F-8 & 64.4 & 62.2 & 52.0 & 48.0 \\
DeformF2F-8-FT & 64.8 & 62.5 & \textbf{52.4} & \textbf{48.3} \\
\midrule
DeformF2F-8-FT (2 samples per seq.) & 65.5 & 63.8 & \textbf{53.6} & \textbf{49.9} \\
\bottomrule
\end{tabular}
}
\end{center}
\label{tab:mainres}
\end{table*}
Poor results of copy-last-segmentation
reflect the difficulty of the forecasting task.
Our method DeformF2F-8 
outperforms Luc F2F for 4.6 pp mIoU.
In comparison with the state-of-the-art,
we achieve the best mid-term performance,
while coming close to \cite{terwilliger2018recurrent} 
in short-term,
despite a weaker oracle 
(72.5 vs 74.3 mIoU)
and not using optical flow.
Cross entropy fine-tuning
improves results by 0.4 pp mIoU 
both for the short-term
and the mid-term model.
We applied DeformF2F-8-FT to Cityscapes test
and achieved results similar
to those on the validation set:
64.3 mIoU (short-term)
and 52.6 mIoU (mid-term).

The last result in the table shows
benefits of training on more data.
Here we train our F2F model 
on two farthest tuples 
(instead of one)
in each mini-clip.
Cross entropy fine-tuning
is done in the regular way,
since groundtruth is available 
only in the 19th frame in each mini-clip.
We notice significant improvement 
of 0.7 and 1.2 pp mIoU for short-term and
mid-term forecast respectively.

\subsection{Single-Step vs. Autoregressive Mid-term Forecast}
\label{sec:autoreg}
There are two possible options for predicting further than one step into the future: 
i) train a separate single-step model 
   for each desired forecast interval, 
ii) train only one model 
   and apply it autoregressively.
Autoregressive forecast 
applies the same model
in the recurrent manner, 
by using the current prediction 
as input to each new iteration.
Once the model is trained, 
the autoregression can be used 
to forecast arbitrary number 
of periods into the future.
Unfortunately, auto-regression accumulates 
prediction errors from intermediate forecasts.
Hence, the compound forecast tends to be worse 
than in the single-step case.
\begin{table*}[h]
\caption{Validation of auto-regressive 
  mid-term forecast on Cityscapes val.}
\begin{center}
{\small
\setlength{\tabcolsep}{6pt}
\begin{tabular}{lcc}
\toprule
& \multicolumn{2}{c}{Mid-term} \\
DeformF2F-8 variant & mIoU & mIoU-MO \\
\midrule
single-step & \textbf{52.4} & \textbf{48.3} \\ 
autoregressive 3$\times$ & 48.7 & 43.5 \\ 
autoregressive 3$\times$ fine-tuned & 51.2 & 46.5 \\
\bottomrule
\end{tabular}
}
\end{center}
\label{tab:autoreg}
\end{table*}

Table \ref{tab:autoreg} validates
autoregressive models.
The first row shows 
our single-step model (cf.~Table \ref{tab:mainres})
for mid-term forecast.
The middle row shows 
the baseline autoregressive forecast
with our corresponding short-term model.
The last row shows improvement 
due to recurrent fine-tuning
for mid-term prediction, 
while initializing with 
the same short-term model 
as in the middle row.
Fine-tuning brings 2.5pp mIoU improvement
with respect to the autoregressive baseline.
Nevertheless, the single-step model 
outperforms the best autoregressive model.

\begin{table}[b]
  \renewcommand{\arraystretch}{1.3}
  \caption{Single-step and autoregressive per-class results on Cityscapes val.
     Rows denoted with $^\dag$ are evaluated
     only on Frankfurt sequences 
     where long clips are available.
     }
  \label{tab:per_class_iou}
  \tiny
  \centering\medskip
  \resizebox{\textwidth}{!}{%
  \begin{tabular}{c|ccccccccccccccccccc|c}
     & \rotatebox{90}{road} & \rotatebox{90}{sidewalk} & \rotatebox{90}{building} & \rotatebox{90}{wall} & \rotatebox{90}{fence} & \rotatebox{90}{pole} & \rotatebox{90}{traffic light} & \rotatebox{90}{traffic sign} & \rotatebox{90}{vegetation} & \rotatebox{90}{terrain} & \rotatebox{90}{sky} & \rotatebox{90}{person} & \rotatebox{90}{rider} & \rotatebox{90}{car} & \rotatebox{90}{truck} & \rotatebox{90}{bus} & \rotatebox{90}{train} & \rotatebox{90}{motorcycle} & \rotatebox{90}{bicycle} & mean \\
    \toprule
    Oracle & 97.5 & 81.6 & 90.7 & 50.1 & 53.4 & 56.1 & 60.3 & 70.8 & 90.9 & 60.9 & 92.9 & 75.9 & 53.0 & 93.2 & 67.4 & 84.4 & 72.0 & 54.5 & 71.7 & 72.5 \\
    \hline
    Short-term & 96.1 & 73.9 & 87.0 & 47.9 & 50.8 & 35.8 & 51.4 & 57.2 & 86.7 & 56.0 & 88.7 & 58.8 & 41.4 & 86.3 & 64.8 & 75.2 & 63.7 & 48.5 & 60.6 & 64.8 \\
    Mid-term & 93.2 & 61.2 & 79.6 & 41.6 & 45.1 & 15.1 & 31.9 & 33.2 & 78.3 & 49.1 & 80.1 & 39.1 & 24.6 & 72.9 & 60.0 & 63.5 & 46.5 & 37.5 & 41.9 & 52.4 \\
    \hline
    AR-3$^\dag$ & 95.8 & 71.1 & 84.9 & 42.0 & 52.2 & 35.0 & 46.2 & 53.5 & 85.0 & 50.0 & 88.0 & 59.0 & 36.6 & 86.2 & 68.5 & 71.7 & 60.6 & 51.8 & 58.0 & 63.0 \\
    AR-6$^\dag$ & 94.3 & 64.2 & 80.9 & 37.6 & 48.6 & 23.5 & 35.4 & 40.6 & 80.1 & 46.8 & 82.8 & 48.4 & 26.3 & 78.8 & 64.9 & 66.0 & 50.0 & 44.5 & 49.4 & 56.0 \\
    AR-9$^\dag$ & 93.4 & 61.1 & 78.0 & 37.7 & 46.2 & 17.5 & 28.4 & 30.9 & 77.0 & 44.5 & 79.3 & 41.8 & 23.2 & 74.4 & 63.7 & 60.7 & 34.0 & 42.1 & 43.5 & 51.5 \\
    AR-12$^\dag$ & 92.6 & 57.7 & 75.3 & 36.5 & 44.1 & 13.5 & 21.5 & 25.4 & 74.2 & 42.2 & 75.7 & 35.5 & 18.3 & 69.8 & 57.1 & 53.8 & 29.6 & 37.7 & 37.3 & 47.3 \\
    AR-15$^\dag$ & 91.6 & 53.8 & 72.9 & 35.7 & 42.0 & 10.8 & 17.9 & 20.1 & 71.1 & 36.4 & 71.6 & 31.6 & 13.2 & 64.5 & 40.6 & 48.0 & 34.7 & 24.4 & 32.9 & 42.9 \\
    AR-18$^\dag$ & 90.7 & 51.4 & 71.0 & 33.9 & 40.9 & 09.1 & 14.7 & 15.6 & 68.9 & 34.5 & 69.0 & 29.2 & 12.4 & 60.4 & 38.2 & 46.6 & 16.8 & 25.1 & 28.2 & 39.9 \\
    \bottomrule
  \end{tabular}
  }
\end{table}
Table~\ref{tab:per_class_iou} shows 
per-class auto-regressive performance
for different forecasting offsets. 
The three sections correspond to the oracle, 
two single-step models,
and autoregressive application
of the last model from  
Table \ref{tab:autoreg}.
Autoregressive experiments have been performed
on 267 sequences from the 
Frankfurt subset of Cityscapes val. 
Long clips are not available for other cities. 

The performance drop due to forecasting
is largest for class person
among all of moving object classes.
We believe that this is 
because persons are articulated:
it is not enough for the model 
to determine the new position
of the object center,
the model also needs to determine
positions and poses 
of the parts (legs and arms).
Poles seem to be 
the hardest static class
because of their thin shape.
Qualitative results 
(e.g. last two rows of fig.~\ref{fig:midterm})
show that pole often gets dominated
by large surrounding classes
(building, sidewalk, road etc.).

\begin{figure}[h]
  \centering
  \includegraphics[width=0.8\textwidth]
    {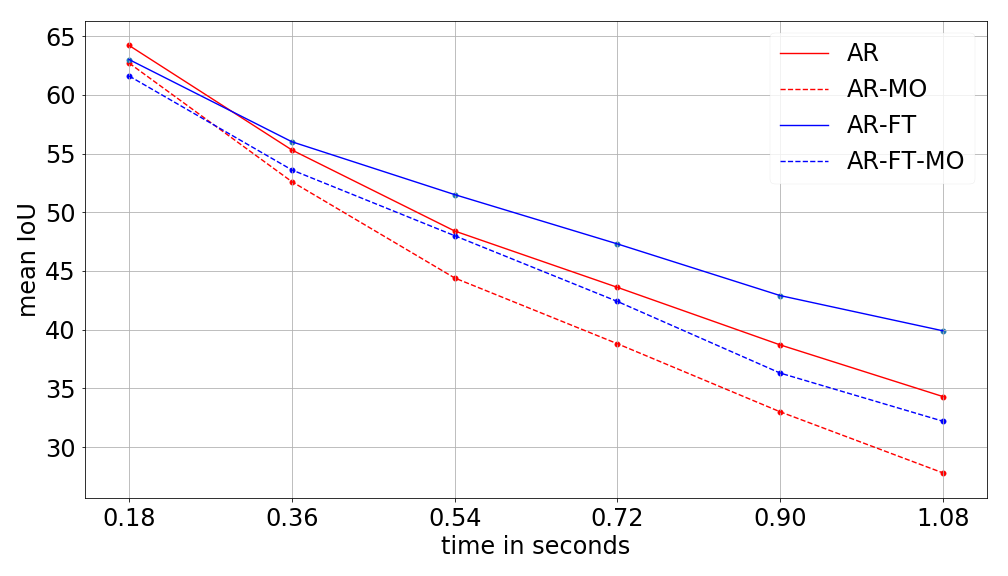}
  \caption{Autoregressive mIoU performance
    at different forecasting offsets
    for the straight 
    short-term model (red) 
    and the model fine-tuned 
    for mid-term prediction (blue).}
    \label{fig:miou_per_t}
\end{figure}

Figure~\ref{fig:miou_per_t} plots mIoU results
from the third section of Table~\ref{tab:per_class_iou}
for various temporal offsets of the future frame,
and explores contribution of autoregressive fine-tuning.
We show mIoU and mIoU-MO 
(solid and dashed lines resp.) 
for a straight autoregressive model (red), 
and a model that was autoregressively fine-tuned for mid-term forecast (blue).

\subsection{Validation of Deformable Convolutions}

Table \ref{tab:deformvsreg} compares 
the mIoU performance and 
the number of parameters
for various design choices. 
Our DeformF2F-5 model achieves 
a 4-fold decrease in the number of parameters 
with respect to Luc F2F. 
Dilated and deformable convolutions 
achieve the largest impact 
in mid-term forecasting 
where the feature displacements 
are comparatively large. 
Dilation achieves a slight improvement
on mid-term prediction. 
Deformable convolutions improve both 
the short-term and mid-term results 
while significantly outperforming 
the dilation models.
This clearly validates the choice 
of deformable convolutions 
for F2F forecasting.
\begin{table*}[h]
\caption{Validation of plain, 2$\times$ dilated 
  and deformable 
  convolutions 
  on Cityscapes.}
\begin{center}
{\small
\setlength{\tabcolsep}{6pt}
\begin{tabular}{lccccc}
\toprule
& \multicolumn{2}{c}{Short-term} & \multicolumn{2}{c}{Mid-term} \\
& mIoU & mIoU-MO & mIoU & mIoU-MO & \#params \\
\midrule
Luc F2F & 59.8 & 56.7 & 45.6 & 39.0 & 5.50M \\
ConvF2F-5 & 60.4 & 56.6 & 43.8 & 36.3 & 1.30M \\
DilatedF2F-5 & 60.0 & 56.9 & 45.6 & 38.8 & 1.30M \\
DeformF2F-5 & \textbf{63.4} & \textbf{61.5} & \textbf{50.9} & \textbf{46.4} & 1.43M \\

\bottomrule
\end{tabular}
}
\end{center}
\label{tab:deformvsreg}
\end{table*}

\subsection{Ablation of the Number of Input Frames}

Table \ref{tab:frames_abl} investigates 
the impact of the number of input frames 
to short-term and mid-term performance. 
We always sample frames three steps apart.
For instance, the second row in the table
observes frames at t-6, t-3, and t.
The model operating on a single frame 
performs significantly worse 
than the models which observe multiple frames.
Such model can only predict the movement direction
from object posture and/or orientation, 
while it is often very hard
to forecast the magnitude of motion 
without looking at least one frame in the past.
Models operating on two and three frames 
produce comparable short-term forecast 
with respect to the four frame model.
Adding more frames from the past
always improves the accuracy 
of mid-term forecasts.
This suggests that the models 
benefit from past occurrences
of the parts of the scene
which are disoccluded 
in the forecasted frame.
This effect is visible 
only in mid-term prediction,
since such occlusion-disocclusion patterns 
are unlikely to occur
across short time intervals.

\begin{table*}[h]
\caption{Ablation 
  of the number of input frames. 
  Two input frames are enough 
  for short-term forecasting. 
  More input frames improve performance 
  of mid-term forecasts.
}
\begin{center}
{\small
\setlength{\tabcolsep}{6pt}
\begin{tabular}{lccccc}
\toprule
&& \multicolumn{2}{c}{Short-term} & \multicolumn{2}{c}{Mid-term} \\
& \#frames & mIoU & mIoU-MO & mIoU & mIoU-MO \\
\midrule
\multirow{4}{*}{DeformF2F-8}
  & 4 & 64.4 & 62.2 
  & \textbf{52.0} & \textbf{48.0} \\
  & 3 & 64.4 & 62.5 & 50.9 & 46.2 \\
  & 2 & \textbf{64.5} & \textbf{62.6} 
  & 50.7 & 46.2 \\
  & 1 & 57.7 & 54.3 & 44.2 & 37.8 \\
\bottomrule
\end{tabular}
}
\end{center}
\label{tab:frames_abl}
\end{table*}

\subsection{Could a Forecast Improve 
the Prediction in the Current Frame?}

We consider an ensemble of a single-frame model 
which observes the current frame
and a forecasting model 
which observes past frames.
The predictions of the ensemble
are a weighted average of
softmax activations 
of the two models:
\begin{equation}
    P(\mathbf{Y}_{t+3}| 
        \mathbf{X}_{t-9}, .., \mathbf{X}_{t+3}) 
    = \lambda \cdot 
      P(\mathbf{Y}_{t+3}| \mathbf{X}_{t+3}) +
    (1 - \lambda) \cdot 
    P(\mathbf{Y}_{t+3}| 
        \mathbf{X}_{t-9}, .., \mathbf{X}_{t}) 
    \;.
    \label{eq:ens}
\end{equation}
Similar results are achieved
for $\lambda \in [0.7, 0.9]$. 
Table~\ref{tab:ensemble} presents 
experiments on City\-scapes val. 
The first two rows show the oracle
and our best short-term model.
The third row ensembles 
the previous two models 
according to (\ref{eq:ens}).
We observe 0.3pp improvement 
over the single-frame model. 
This may be interesting 
in autonomous driving applications 
which
would need 
semantic segmentation
for the current 
and the future frame
in each time instant.
In that case,
the proposed ensemble
would require no additional cost,
since the forecast 
from the previous time instant
can be cached.
On the other hand,
evaluating an ensemble
of two single-frame models
would imply double 
computational complexity.

\begin{table*}[h]
\caption{Performance of the ensemble 
  of a single-frame model
  which observes the current frame
  with a forecasting model which observes 
  only the four past frames.
}
\begin{center}
{\small
\setlength{\tabcolsep}{6pt}
\begin{tabular}{lcc}
\toprule
& mIoU & mIoU-MO \\
\midrule
Single frame model & 72.5 & 71.5 \\
DeformF2F-8-FT & 64.8 & 62.5 \\ 
Ensemble & \textbf{72.8} & \textbf{71.8} \\ 
\bottomrule
\end{tabular}
}
\end{center}
\label{tab:ensemble}
\end{table*}

\subsection{Qualitative Results}

\begin{figure}[b]
    \centering
    \includegraphics[width=\textwidth]{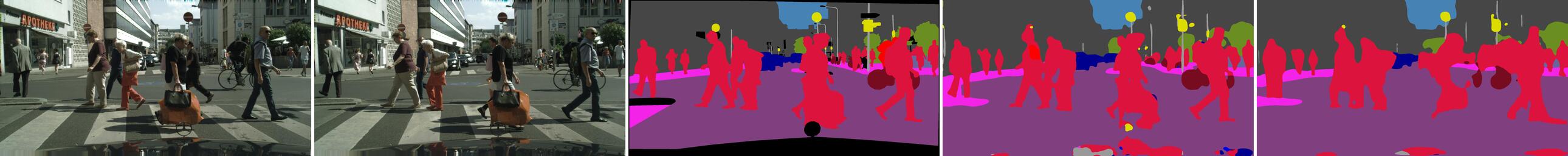}
    \includegraphics[width=\textwidth]{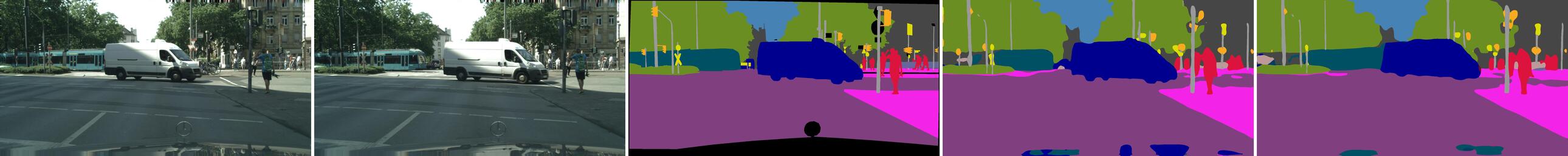}
    \includegraphics[width=\textwidth]{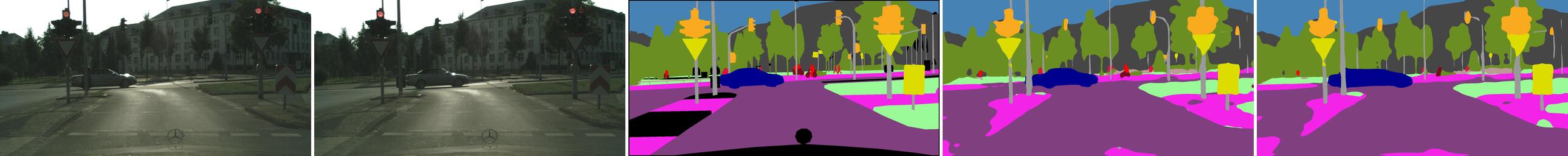}
    \caption{Short-term semantic segmentation forecasts (0.18\,s into the future) for 3 sequences. 
    The columns show 
    i) the last observed frame,
    ii) the future frame, 
    iii) the groundtruth segmentation, 
    iv) our oracle, and 
    v) our semantic segmentation forecast.
    }
    \label{fig:shortterm}
\end{figure}

\begin{figure}[h]
    \centering
    \includegraphics[width=\textwidth]{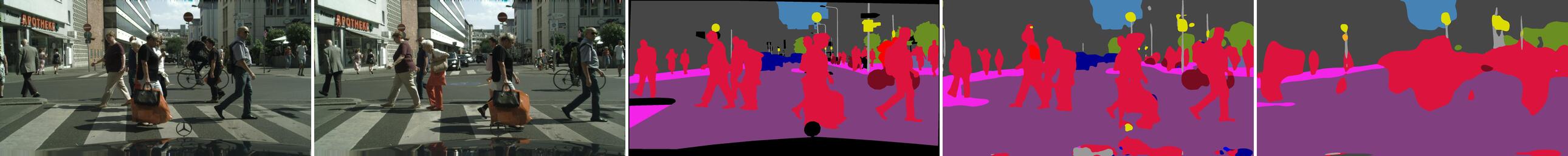}
    \includegraphics[width=\textwidth]{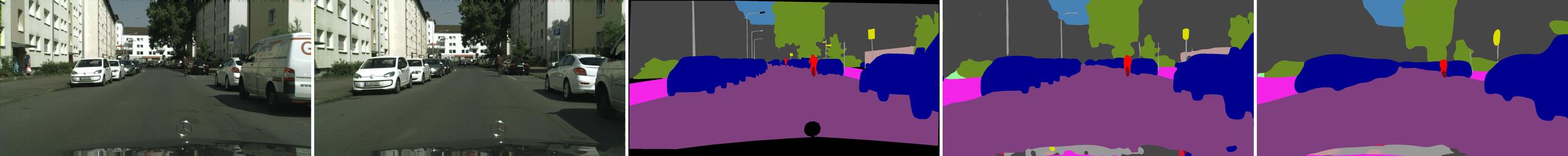}
    \includegraphics[width=\textwidth]{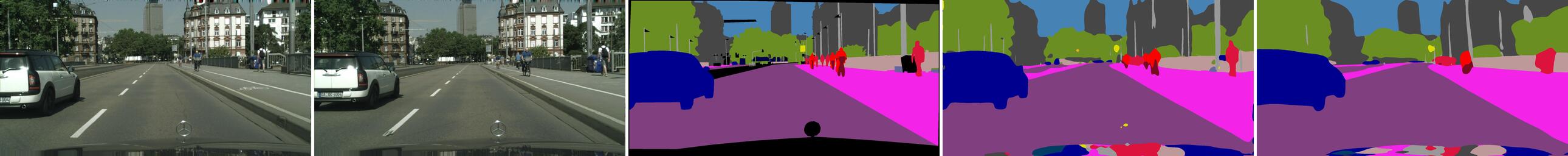}
    \includegraphics[width=\textwidth]{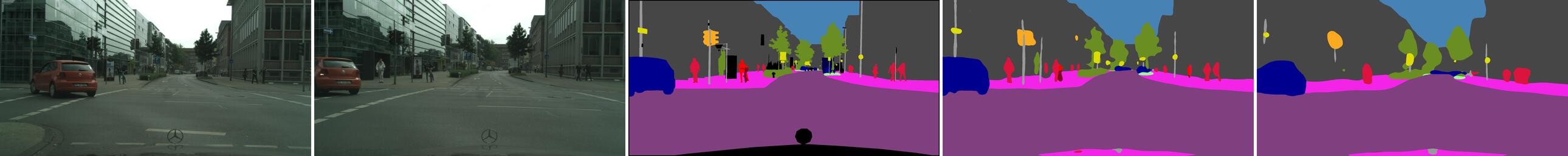}
    \includegraphics[width=\textwidth]{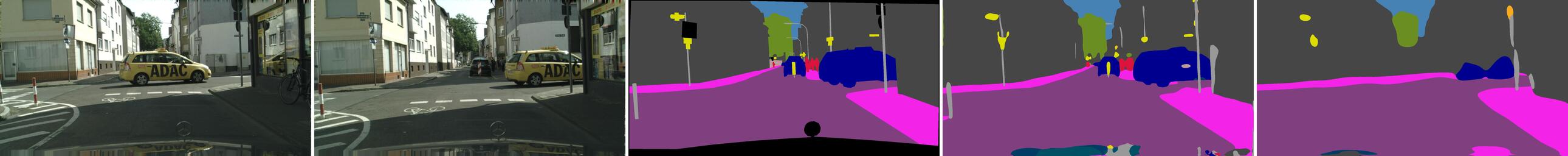}
    \caption{Mid-term semantic segmentation predictions (0.5 s into the future) 
    for 5 sequences. 
    The columns show 
    i) the last observed frame,
    ii) the future frame, 
    iii) the ground truth segmentation, 
    iv) our oracle, and 
    v) our semantic segmentation forecast.
    }
    \label{fig:midterm}
\end{figure}

Figures \ref{fig:shortterm} and \ref{fig:midterm} show forecasted semantic segmentation on Cityscapes val for short-term and mid-term predictions respectively. 
We observe loss of spatial detail when 
forecasting sequences with greater dynamics 
and when predicting further into the future 
(cf.\ the first row in figures \ref{fig:shortterm} and \ref{fig:midterm}).
The row 4 in figure \ref{fig:midterm} shows
a red car turning left. 
%
Our model inferred the future 
spatial location of the car 
quite accurately. 
The last row shows a car which 
disoccludes the road opposite the camera.
Our model correctly inferred the car motion
and in-painted the disoccluded scenery
in a feasible although not completely correct manner.

\begin{figure}[h]
  \centering
  \includegraphics[width=\textwidth]{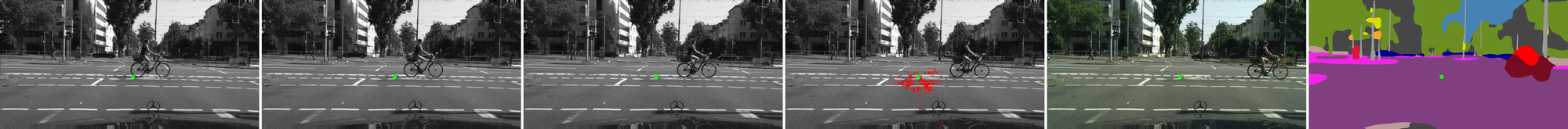}
  \includegraphics[width=\textwidth]{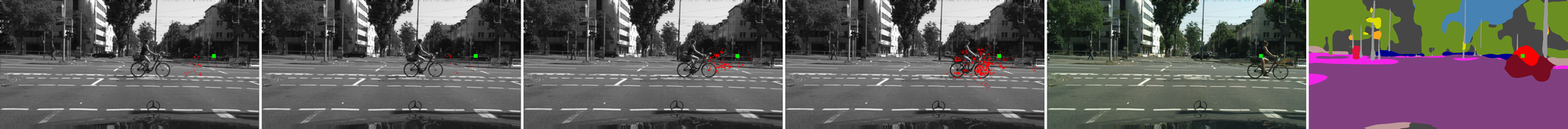}
  \includegraphics[width=\textwidth]{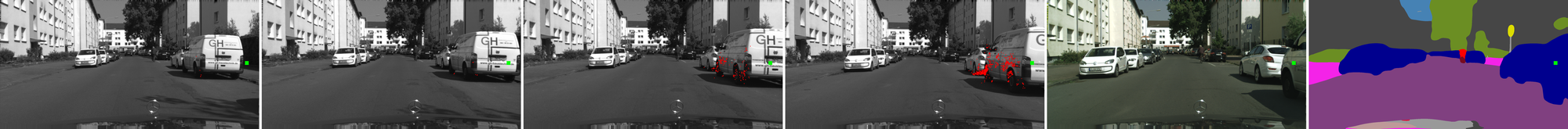}
  \includegraphics[width=\textwidth]
  {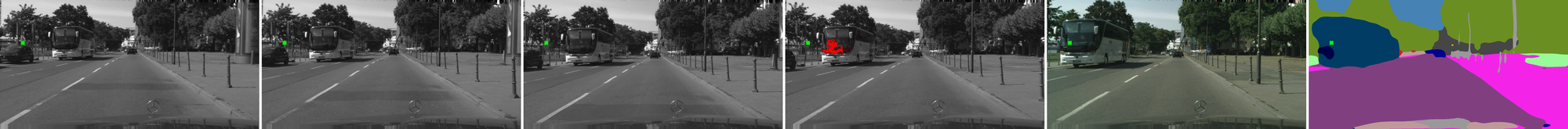}
  \caption{Effective receptive field 
    of mid-term forecast in 4 sequences. 
    Columns show the four input frames,
    the future frame 
    t+9
    and the corresponding
    semantic segmentation forecast. 
    We show pixels with the strongest
    gradient of log-max-softmax (red dots)
    in a hand-picked pixel (green dot) 
    w.r.t. the each of the input frames. 
  }
  \label{fig:gradients}
\end{figure}

\paragraph{Effective receptive field.}
We express the effective receptive field
by measuring partial derivation of
log-max-softmax \cite{luo2016understanding}
with respect to the four input images.
The absolute magnitude of these gradients 
quantifies the importance 
of particular pixels
for the given prediction.
Figure~\ref{fig:gradients} 
visualizes the results
for our DeformF2F-8-FT
mid-term model.
The four leftmost columns 
show input images,
while the two rightmost columns show 
the future image
(unavailable to the model),
and the semantic forecast.
The green dot in the two rightmost
columns designates the examined prediction.
The red dots designate pixels in which 
the absolute magnitude of the gradient 
of the examined prediction 
is larger than a threshold. 
The threshold is dynamically set 
to the value of the k-th 
(k = 3000, top 0.15 percent)
largest gradient within 
the last observed frame (t). 
In other words, we show pixels with
top k gradients in the last observed frame,
as well as a selection of pixels
from the other frames
according to the same threshold.
We notice that most important pixels 
come from the last observed frame.
Row 1 considers a static pixel
which does not generate strong gradients 
in frames t-3, t-6, and t-9.
Other rows consider dynamic pixels.
We observe that the most important pixels 
for a given prediction
usually correspond 
to object location in past frames.
Distances between object locations 
in the last observed and the forecasted frame
are often larger than 300 pixels.
This emphasizes the role of deformable convolutions
since the F2F model with plain convolutions
is unable to compensate for such large offsets.
The figure also illustrates the difficulty 
of forecasting in road-driving videos,
and the difference of this task
with respect to single-frame semantic segmentation.
These visualizations allow us 
to explain and interpret 
successes and failures of our model
and to gauge the range of its predictions.
In particular we notice that 
most mid-term decisions rely only 
on pixels from the last two frames.
This is in accordance 
with mid-term experiments 
from Table \ref{tab:frames_abl} 
which show that frames t-6 and t-9
contribute only 1.3pp mIoU.

\section{Conclusion and Future Work}

We have presented a novel method 
for anticipating 
semantic segmentation of 
future frames in driving scenarios 
based on feature-to-feature (F2F) 
forecasting.
Unlike previous methods, 
we forecast the most abstract 
backbone features
with a single F2F model.
This greatly improves the inference speed
and favors the forecasting performance
due to coarse resolution
and high semantic content
of the involved features.
The proposed F2F model
is based on deformable convolutions
in order to account for geometric nature
of F2F forecasting.
We use a lightweight single-frame model
without lateral connections,
which allows to adapt the upsampling path
to F2F noise by fine-tuning
with respect to groundtruth labels.
We perform experiments on 
the Cityscapes dataset.
To the best of our knowledge, 
our mid-term semantic segmentation forecasts 
outperform all previous approaches.
Our short-term model 
comes second only to a method
which uses a stronger single-frame model
and relies on optical flow.
Evaluation on Cityscapes test
suggests that our validation performance 
contains very little bias (if any).
Suitable directions for future work include
adversarial training of the upsampling path,
complementing image frames with optical flow,
investigating end-to-end learning,
as well as evaluating performance
on the instance segmentation task.

\section*{Acknowledgment}
This work has been funded by Rimac Automobili.
This work has been partially supported
by European Regional Development Fund
(DATACROSS) under grant KK.01.1.1.01.0009.
We thank Pauline Luc and Jakob Verbeek
for useful discussions 
during early stages of this work.
\clearpage
\bibliographystyle{splncs04}
\bibliography{egbib}

\end{document}


\pagestyle{headings}
	\mainmatter

	\title{Single Level Feature-to-Feature Forecasting
    \texorpdfstring{\\}{}
    with Deformable Convolutions
    \texorpdfstring{\\}{}
    - Supplementary Material}

	\titlerunning{Supplementary Material}
	\authorrunning{}
\author{
Josip {\v{S}}ari{\'c}\inst{1} \and
Marin Or{\v{s}}i{\'c}\inst{1} \and
Ton{\'c}i Antunovi{\'c}\inst{2} \and
Sacha Vra{\v{z}}i{\'c}\inst{2} \and
Sini{\v{s}}a {\v{S}}egvi{\'c}\inst{1}
}

\institute{
$^1$ University of Zagreb, Faculty of Electrical Engineering and Computing, Croatia\\
$^2$ Rimac Automobili, Sveta Nedelja, Croatia\\
}

	\maketitle
	
  \section{Additional Mid-Term Results of the Standard Model}
  \subsection{Qualitative Discussion of Interesting Cases}
\begin{figure}[h]
    \centering
    \includegraphics[width=\textwidth]{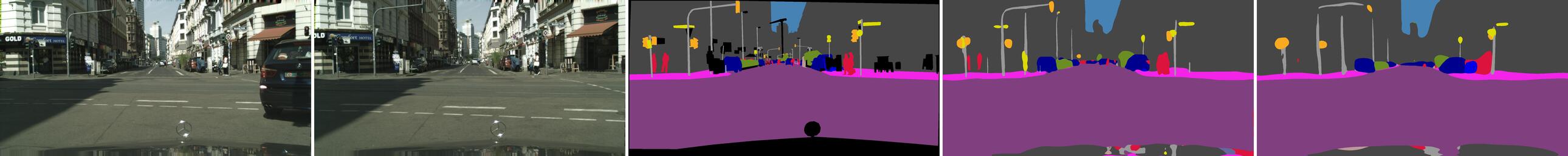}
    \includegraphics[width=\textwidth]{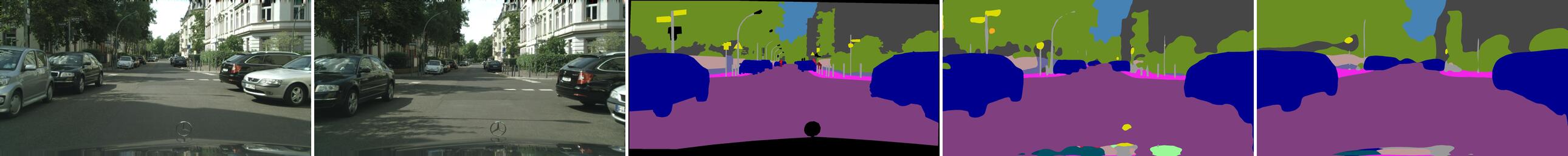}
    \includegraphics[width=\textwidth]{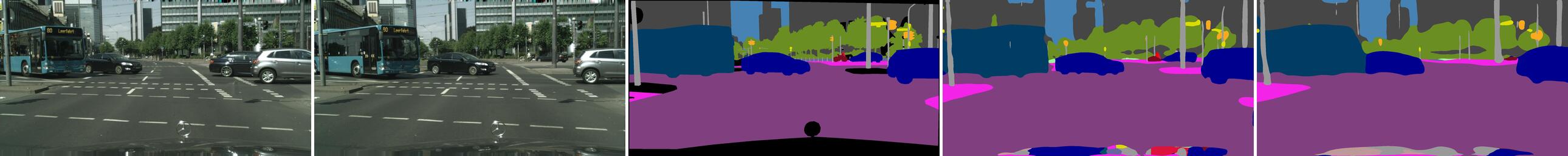}
    \includegraphics[width=\textwidth]{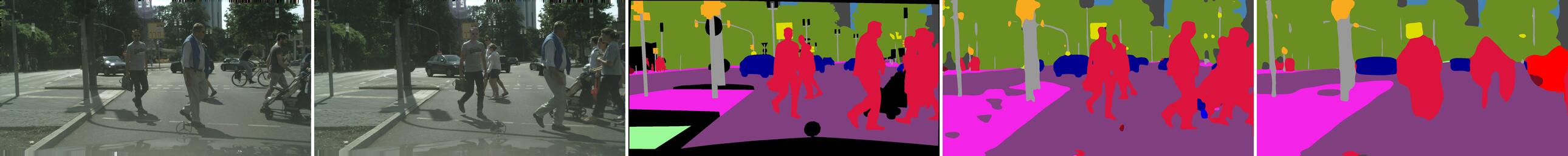}
    \includegraphics[width=\textwidth]{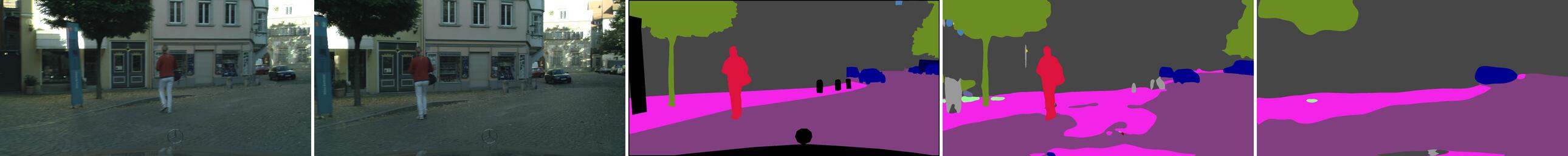}
    \includegraphics[width=\textwidth]{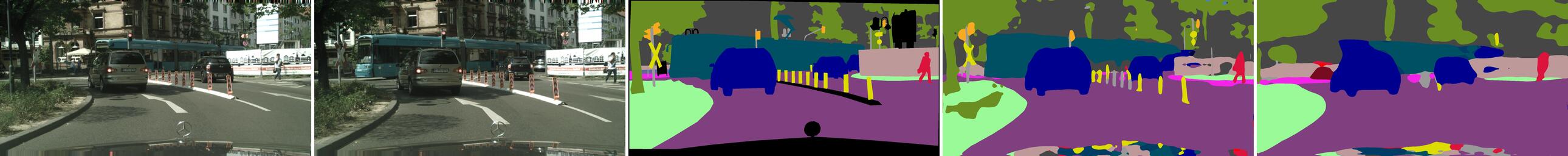}
    \caption{
        Mid-term semantic segmentation predictions (0.54 s into the future) 
        for 5 sequences. 
        The columns show 
        i) the last observed frame,
        ii) the future frame, 
        iii) the ground truth segmentation, 
        iv) our oracle, and 
        v) our semantic segmentation forecast.
    }
    \label{fig:midterm}
    \end{figure}

    Figure~\ref{fig:midterm} shows additional qualitative results 
  obtained by our mid-term model 
  based on ResNet-18 and DeformF2F-8.
    Row 1 showcases the ability of inpainting 
  by considering a wider context.
  A car on the right is making the turn
  and unoccludes the part of the future frame
  which has not been visible in previous frames.
  The model correctly reconstructs the scene
  by forecasting a feasible configuration
  of road, sidewalk and building.
  Row 2 shows a situation where 
  a car on the left leaves the scene,
  while its place is taken by another car.
  The model predicts the dynamics correctly,
  and forecasts the future frame with only one car.
  Row 3 shows two cars with different speeds
  in the right part of the scene.
  The faster car gets occluded 
  by the slower car in the future frame.
  Our model appears to understand this relationship
  and succeeds to forecast the front car quite precisely.
  Row 4 shows a very dynamic scene with 
  several articulated objects.
  Our model is unable to recover details 
  which are present in the oracle prediction,
  and incorrectly infers that the cyclist
  got apparated in front of the pedestrians.
  Row 5 shows a simple scene with one pedestrian
  moving pretty slowly.
  Our model fails to segment the pedestrian,
  although the oracle predicts it correctly.
  Row 6 shows a scene where a train
  is moving in front of the static car.
  Our model fails to infer the correct 
  future location of the train,
  and at some pixels forecasts a wrong class
  (car instead of train).
  We also observe loss of small objects
  such as traffic signs and poles next to the road.
  
  \subsection{Qualitative Comparison of 
  Deformable vs Dilated Convolutions}
\begin{figure}[h]
    \centering
    \begin{tabular}{c}
        \includegraphics[width=\textwidth]{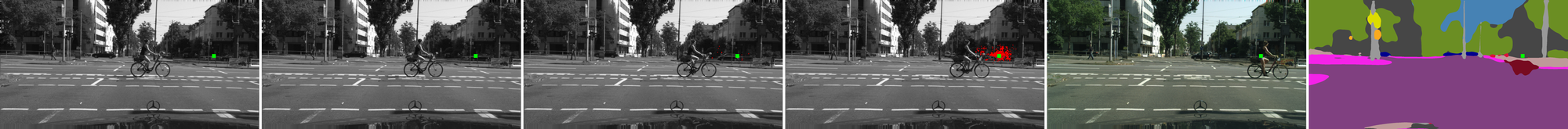} \\
        \includegraphics[width=\textwidth]{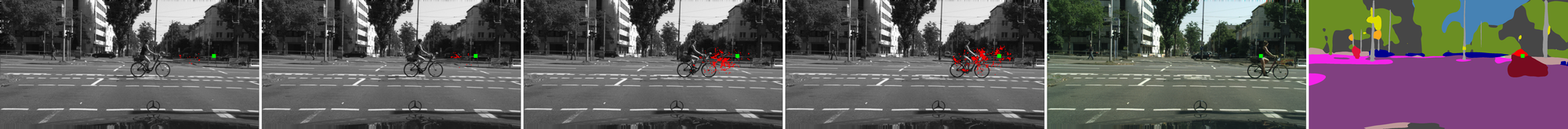} \\
        (a) \\
        \includegraphics[width=\textwidth]{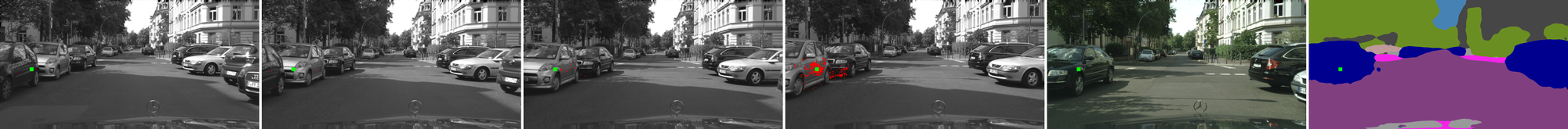} \\
        \includegraphics[width=\textwidth]{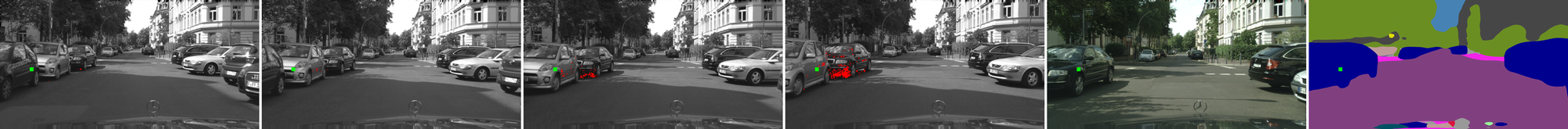} \\
        (b)
    \end{tabular}
    \caption{
        Comparison of effective receptive fields for F2F models
        with dilated (top) and deformable (bottom) convolutions
        on two mid-term sequences.  
        Columns show the four input frames,
        the future frame 
        t+9
        and the corresponding
        semantic segmentation forecast. 
        We show pixels with the strongest
        gradient of log-max-softmax (red dots)
        in a hand-picked pixel (green dot) 
        w.r.t. the each of the input frames. 
    }
    \label{fig:dilvsdeform}
    \end{figure}

    As we showed in the paper (cf.\ Table 4) 
  models with deformable convolutions 
  outperform their dilation counterparts
  for 5 pp mIoU when forecasting nine timesteps ahead.
  We believe the difference is caused by 
  learnable and adaptive grid sampling locations
  in deformable convolutions.
  Figure~\ref{fig:dilvsdeform} compares
  effective receptive fields (details are explained in the paper, cf.\ Figure~5)
  for models
  DilatedF2F-5 and DeformF2F-5,
  which are based on dilated and 
  deformable convolutions respectively.
  The situation (a) features a cyclist 
  which moves across the scene 
  from left to the right.
  We observe that the gradients of the dilated model 
  follow the regular grid layout 
  and are unable to reach the cyclist.
  On the other hand, the gradients of the deformable model 
  are noticeably displaced towards the cyclist.
  The situation (b) features two cars 
  in the front left part of the scene.
  The car in front is about to leave the current frame 
  due to ego-motion of the camera.
  The dilated model incorrectly relies 
  on the pixels of the leaving car,
  while the gradients of the deformable model 
  clearly correspond to the correct car.

\subsection{Spatial Layout of Forecast Uncertainty}

\begin{figure}[h]
  \centering
  \includegraphics[width=\textwidth]{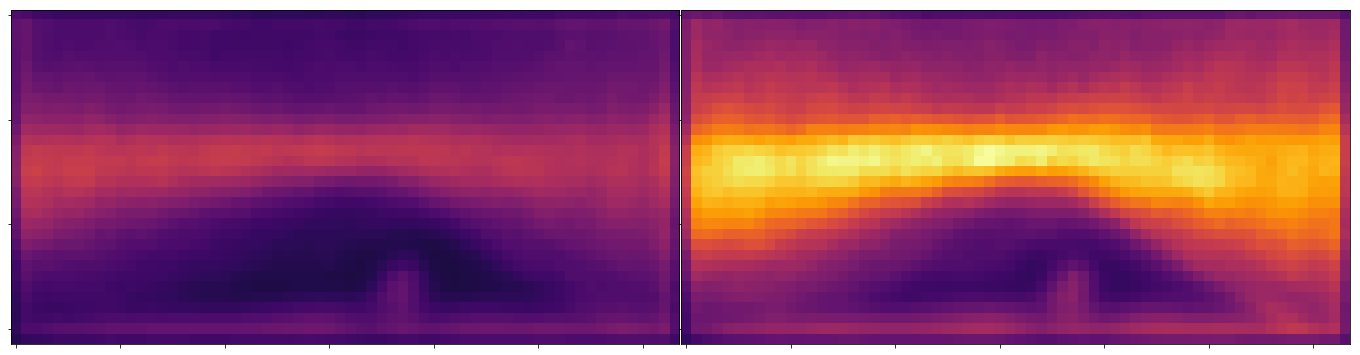}
  \caption{Distribution of 
    the average mean-square-error 
    of F2F predictions on Cityscapes val. Darker colors correspond to lower values. Largest errors occur around the horizon.
  }
    \label{fig:mse}
\end{figure}

We visualize the forecast uncertainty layout  
as a spatial map of average forecast errors
across the whole Cityscapes validation subset.
We quantify the forecast error 
as a mean-square-error 
of F2F predictions across all feature maps.
Figure \ref{fig:mse} shows 
the resulting maps of forecast uncertainty.
The two sub-figures correspond 
to short-term (left) and 
mid-term predictions (right).
Both maps are shown on the same scale,
while the error increases 
from darker to lighter colors.
We observe that largest errors 
occur around the horizon, 
because most of dynamics 
in Cityscapes scenes
happens right there.
The shapes of the two error distributions 
are very similar, 
while the magnitude is significantly larger
for mid-term predictions.

  \section{Increasing the Capacity of the Single-frame Model}
  
  We investigate the influence of 
  the single-frame model capacity
  on the accuracy of the
  semantic segmentation forecast.
  We conduct short-term experiments
  with our standard F2F model and 
  a stronger single-frame model.
  The differences with respect to 
  our standard single-frame model
  from the paper are as follows.
  First, the new single-frame model
  has a different backbone 
  (ResNet-18 vs. DenseNet-121).
  It is important to notice that DenseNet-121
  has 1024 feature maps at the output 
  of the last processing block.
  This is a two-fold increase 
  with respect to the ResNet-18.
  Consequently, the number of F2F parameters
  increases from 1.9M to 2.8M.
  Second, the new single-frame model 
  has a wider SPP module 
  and a wider first module
  in the upsampling path 
  (256 vs 128 channels).
  The F2F training scheme is the same
  as described in the paper,
  except that we normalize DenseNet features
  with per-channel mean and standard deviation
  across the whole training set.
  This procedure is not neccessary for ResNet-18,
  because its last convolutional unit
  has a built-in batchnorm layer.

  \begin{table*}[h]
\caption{Segmentation forecasting on the Cityscapes validation set.}
\begin{center}
{\small
\setlength{\tabcolsep}{6pt}
\begin{tabular}{lcccc}
\toprule
& \multicolumn{2}{c}{ResNet-18} & \multicolumn{2}{c}{DenseNet-121} \\
& mIoU & mIoU-MO & mIoU & mIoU-MO \\
\midrule
Oracle & 72.5 & 71.5 & 74.9 & 74.1 \\
Copy last segmentation & 52.2 & 48.3 & 53.0 & 48.7 \\
\midrule
DeformF2F-8 & 64.4 & 62.2 & 65.3 & 62.6 \\
DeformF2F-8-FT & 64.8 & 62.5 & 65.9 & 63.2  \\
\bottomrule
\end{tabular}
}
\end{center}
\label{tab:mainres}
\end{table*}
  
  The DenseNet-based single-frame model 
  outperforms its ResNet counterpart
  for 2.4 pp mIoU on Cityscapes val.
  The copy-last-segmentation baseline
  benefits less while nevertheless improving 
  0.8 pp mIoU and 0.4 pp mIoU-MO.
  Semantic forecasting with 
  the DeformF2F-8 model improves for
  0.9 pp mIoU and 0.4 pp mIoU-MO,
  while the fine-tuned variant improves for 
  1.1 pp mIoU and 0.7 pp mIoU-MO.
  We observe that the difference
  between the two single-frame models 
  is greater than the difference 
  at the semantic forecasting level.